\documentclass[letterpaper, 10 pt, conference]{ieeeconf}  %

\IEEEoverridecommandlockouts                              %

\usepackage{cite}
\usepackage{amsmath,amssymb,amsfonts}
\usepackage{algorithmic}
\usepackage{graphicx}
\usepackage{textcomp}
\usepackage{xcolor} 
\usepackage{url}
\usepackage{subcaption}
\usepackage{booktabs}
\def\BibTeX{{\rm B\kern-.05em{\sc i\kern-.025em b}\kern-.08em
    T\kern-.1667em\lower.7ex\hbox{E}\kern-.125emX}}

\title{\LARGE \bf
Rainbow-DemoRL: Combining Improvements in Demonstration-Augmented Reinforcement Learning
}

\author{Dwait Bhatt$^{1}$, Shih-Chieh Chou$^{2}$, and Nikolay Atanasov$^{1}$%
\thanks{This work was supported by Technology Innovation Program 20018112 (Development of autonomous manipulation and gripping technology using imitation learning based on visual and tactile sensing) funded by the Ministry of Trade, Industry \& Energy (MOTIE), Korea.}%
\thanks{$^{1}$Dwait Bhatt and Nikolay Atanasov are with the Department of Electrical and Computer Engineering, University of California San Diego, La Jolla, CA, USA, e-mails: {\tt\small \{dhbhatt,natanasov\}@ucsd.edu}.}%
\thanks{$^{2}$Shih-Chieh Chou is with the Department of Mechanical and Electro-Mechanical Engineering, National Sun Yat-sen University, Taiwan.}%
}

\begin{document}

\maketitle
\thispagestyle{empty}
\pagestyle{empty}

\begin{abstract}
Several approaches have been proposed to improve the sample efficiency of online reinforcement learning (RL) by leveraging demonstrations collected offline. The offline data can be used directly as transitions to optimize RL objectives, or offline policy and value functions can first be learned from the data and then used for online finetuning or to provide reference actions. While each of these strategies has shown compelling results, it is unclear which method has the most impact on sample efficiency, whether these approaches can be combined, and if there are cumulative benefits. We classify existing demonstration-augmented RL approaches into three categories and perform an extensive empirical study of their strengths, weaknesses, and combinations to isolate the contribution of each strategy and determine effective hybrid combinations for sample-efficient online RL. Our analysis reveals that directly reusing offline data and initializing with behavior cloning consistently outperform more complex offline RL pretraining methods for improving online sample efficiency.
\end{abstract}

\section{Introduction}

Online reinforcement learning (RL) \cite{suttonbarto} has demonstrated impressive performance in game settings \cite{dqn, alphago, alphastar} and, more recently, has shown potential to advance control of physical robot systems \cite{deeprlrealsuccess, reallocomotionrl, realhumanoidrl}.
A key benefit of online RL is that it enables an intelligent system to adapt continuously to changing conditions by learning directly from ongoing interactions. This has long been viewed as having potential to enable continuous lifelong learning in robotics \cite{rewardisenough}. However, two major challenges prevent widespread deployment of online RL on robot systems: (a) low sample-efficiency, which makes learning from scratch on physical hardware prohibitively slow and costly, and (b) the need for random exploration during learning, which can lead to unsafe behaviors when interacting with the physical world.

To address these challenges, a promising direction has been to leverage offline demonstrations to accelerate and improve online RL. Several diverse strategies have been proposed for this purpose. First, some methods use demonstration data directly within the online training loop. This can be done by pre-filling the replay buffer with expert transitions to bootstrap learning from high-quality examples \cite{rlpd} or by adding an auxiliary supervised loss to the RL objective to encourage the agent to mimic the expert's actions \cite{td3bc}. Second, another class of methods uses the data to learn an initial policy or value function via offline RL, which then provides a strong starting point for online finetuning \cite{cql, calql}. Finally, a third approach uses a pre-trained offline policy as a reference for action selection, mixing its outputs with the online policy's actions to guide exploration and improve performance \cite{ibrl, cheq, residualrl}.

While each of these strategies has shown compelling results, they are not mutually exclusive, and many hybrid approaches are possible. This creates a critical knowledge gap: it is unclear which components are complementary and how each one contributes to final performance and learning speed.
The current state of demonstration-augmented RL is reminiscent of the state of deep RL in 2017 when several independent extensions of Deep Q-Networks (DQN) \cite{dqn} were explored in the literature.
Rainbow-DQN \cite{rainbowdqn} systematically combined these extensions and performed an ablation study that clarified the individual importance of each component.

Inspired by this line of work, we conduct a large-scale empirical study of demonstration-augmented online RL with the following objectives:
\begin{itemize}
    \item identify general classes of techniques for demonstration-augmented RL in the relevant prior work,
    
    \item measure the contribution of each strategy to performance and sample efficiency via extensive evaluation,
    
    \item propose effective hybrid combinations of strategies for sample-efficient online learning in robot manipulation.
\end{itemize}

\section{Problem Statement}
Consider a Markov Decision Process (MDP) $\mathcal{M} = (\mathcal{S}, \mathcal{A}, r, p, \gamma)$, where $\mathcal{S} \subseteq \mathbb{R}^n$ and $\mathcal{A} \subseteq \mathbb{R}^m$ are continuous state and action spaces, $r(s_t,a_t)$ and $p(s_{t+1} | s_t, a_t)$ are the reward function and the transition probability density function, where $s_t,s_{t+1} \in \mathcal{S}$, $a_t \in \mathcal{A}$, and $\gamma \in [0,1)$ is the discount factor. The goal in RL is to learn a stochastic policy $\pi(a|s)$ or a deterministic policy $\pi(s)$ to generate actions that maximize the value function $V^\pi(s) = \mathbb{E}_{s_{t+1} \sim p(\cdot | s_t, a_t), a_t \sim \pi(\cdot | s_t)} \left[\sum_t \gamma^t r(s_t,a_t) | s_0=s \right]$. The action value or Q-value of the policy is given by $Q^\pi(s,a) = \mathbb{E}_{s_{t+1} \sim p(\cdot | s_t, a_t), a_t \sim \pi(\cdot | s_t)} [\sum_t \gamma^t r(s_t, a_t) | s_0=s, a_0=a]$. We use $\pi(a| s;\phi)$ or $\pi(s;\phi)$ to denote a stochastic/deterministic policy parameterized by a neural network with weights $\phi$ and $Q^\pi(s,a;\theta)$ to denote its Q-value estimate parameterized by a neural network with weights $\theta$. 

In this paper, we aim to learn a policy $\pi_\text{on}$ and its value function $Q_\text{on}$ by sampling transitions $(s_t,a_t,r_t,s_{t+1})$ from $\mathcal{M}$ online such that $s_t \in \mathcal{S}$, $a_t \sim \pi_{\text{on}}(\cdot | s_t)$, $r_t = r(s_t, a_t)$, $s_{t+1} \sim p(\cdot | s_t, a_t)$. However, in addition, we assume access to an offline dataset of transitions $\mathcal{D}_\text{off} := \{(s_t,a_t,r_t,s_{t+1})\}$ collected by an unknown behavior policy $\pi_\beta$
 such that $s_t \in \mathcal{S}$, $a_t \sim \pi_{\beta}(\cdot | s_t)$, $r_t = r(s_t, a_t)$, $s_{t+1} \sim p(\cdot | s_t, a_t)$. 
From $\mathcal{D}_\text{off}$, we can infer a policy $\pi_\text{off}$, its value function $Q_\text{off}$, or both. These components can be used in several ways to assist learning $\pi_\text{on}$ and $Q_\text{on}$ with few interactions in $\mathcal{M}$, as shown in Sec.~\ref{sec:rdemorl}. Our goal is to minimize the number of online interactions required to learn an optimal $\pi_\text{on}$ by leveraging $\mathcal{D}_\text{off}$, $\pi_\text{off}$ and $Q_\text{off}$.

\section{Preliminaries}
\label{sec:preliminaries}

In this section, we introduce algorithms for training a policy and value function online 
with interactions in $\mathcal{M}$ as well as learning them offline using the dataset $\mathcal{D}_\text{off}$.

\subsection{Online training}
\label{sec:online_rl}

We consider actor-critic algorithms SAC \cite{sac} and TD3 \cite{td3} for online training due to their ability to learn from off-policy transition data, and their success in sample-efficient RL \cite{redq, rlpd}. The training objectives follow policy iteration \cite{suttonbarto} by alternating between optimizing the critic (policy evaluation) and actor (policy improvement) as follows:
\begin{align} 
    L(\theta) &= \mathbb{E}_{(s_t, a_t, \dots, s_{t+h}) \sim \mathcal{D}_\text{on}}
        \left[
            \left(
                Q_\text{on}(s_t,a_t;\theta) - y_t
            \right)^2
        \right]
                \label{eq:critic_loss_general}\\
    L(\phi) &=  -\mathbb{E}_{s \sim \mathcal{D}_\text{on}, a \sim \pi_\text{on}(\cdot | s;\phi)} [Q_\text{on}(s, a)], \label{eq:actor_loss_general}
\end{align}
where $\mathcal{D}_\text{on}$ is a replay buffer which stores transitions $(s_t,a_t,r_t,s_{t+1})$ from online interactions with $\pi_\text{on}(\cdot | s; \phi)$ in $\mathcal{M}$, 
and $y_t$ is the target Q-value which computes the h-step Bellman backup $\mathcal{B}^{\pi_\text{on}} Q_\text{on}$ as follows:
\begin{equation}
    y_t := \sum_{i=0}^{h-1} \gamma^i r_{t+i} + \gamma^h \mathbb{E}_{a' \sim \pi_{\text{on}}(\cdot | s_{t+h}; \phi)} Q_{\text{on}}(s_{t+h}, a'; \bar{\theta})
\label{eq:q_target}
\end{equation}
where $\bar{\theta}$ are target network parameters which are exponential moving averages of the Q-network parameters $\theta$ \cite{ddpg}. SAC and TD3 modify the actor-critic losses in \eqref{eq:critic_loss_general}, \eqref{eq:actor_loss_general} to accommodate entropy and deterministic policies, respectively. Both use Clipped Double Q-learning (CDQ) \cite{td3} to overcome overestimation bias \cite{deepdoubleq}, a situation where states with arbitrarily bad value are estimated as high-value by Q-learning.

To use samples collected online efficiently, it is common to make several gradient updates per interaction, which is referred to as operating in a high update-to-data (UTD) regime. To achieve this, we use the critic ensemble modification of CDQ from REDQ \cite{redq} which calculates target Q-values as:
\begin{equation}
    Q_\text{on}(s',a';\bar{\theta}) = \min_{i \in \mathcal{Z}} Q_{\text{on}, i}(s',a';\bar{\theta}),
\end{equation}
where $\mathcal{Z}$ is a randomly sampled subset of 2 indices from the critic ensemble.

\subsection{Offline training}
\label{sec:offline_training}

Extracting a policy $\pi_{\text{off}}$ from the offline dataset $\mathcal{D}_\text{off}$ is a key objective of offline training, and is typically achieved through imitation learning \cite{imitation, diffusionpolicy}. A simple approach is Behavior Cloning (BC) \cite{bc}, which directly supervises $\pi_{\text{off}}$ with state-action pairs from the dataset. For a stochastic policy, BC minimizes the negative log likelihood of the data, while for a deterministic policy, the objective is to minimize the mean squared error:
\begin{align}
    L(\phi) &= \mathbb{E}_{(s,a) \sim \mathcal{D}_\text{off}}[-\log \pi_\text{off}(a|s; \phi)] &&(\text{stochastic}) \label{eq:bc_stoch}\\
    L(\phi) &= \mathbb{E}_{(s,a) \sim \mathcal{D}_\text{off}}[(\pi_\text{off}(s; \phi) - a)^2] &&(\text{deterministic}) \label{eq:bc_det}
\end{align}

In addition to a policy, a value function $Q_{\text{off}}$ may also be extracted from the dataset. This presents unique challenges compared to standard online learning. The overestimation bias of Q-learning is exacerbated in the offline setting due to the inability to correct estimation errors by observing returns for desired actions. Offline RL methods like CQL \cite{cql} and CalQL \cite{calql} counter this by learning a conservative lower bound on the true Q-function by minimizing Q-values alongside the standard Bellman error objective from \eqref{eq:critic_loss_general}. The CQL critic objective is
\begin{equation} \label{eq:cql}
\begin{aligned} 
    L(\theta) =\ & \mathbb{E}_{s \sim \mathcal{D}_\text{off}, a \sim \pi_\text{off}(\cdot | s;\phi)}
                    \left[
                        Q_\text{off}(s,a;\theta)
                    \right]\\
                & - \mathbb{E}_{(s,a) \sim \mathcal{D}_\text{off}}
                    \left[
                        Q_\text{off}(s,a;\theta)
                    \right] \\
                & + \mathbb{E}_{(s,a,r,s') \sim \mathcal{D}_\text{off}} 
                    \left[ 
                        \left(
                            Q_\text{off}(s,a;\theta) - y(r,s')
                        \right)^2 
                    \right].
\end{aligned}
\end{equation}
While CQL avoids Q-value overestimation, it tends to severely underestimate Q-values. This hinders online finetuning, as the Q-values first undergo a recalibration to the correct scale before effective learning proceeds. CalQL counters this issue by lower bounding the Q-value estimates $Q_\text{off}(s,a)$ by a Monte-Carlo estimate of the value $V^{\pi_\beta}(s)$.
This can be calculated by sampling trajectories of consecutive transitions $\tau=(s_0, a_0, r_0, s_1, \cdots, s_T)$ from $\mathcal{D}_\text{off}$ as follows:
\begin{align}
   V^{\pi_\beta}(s) &\approx \mathbb{E}_{\tau \sim \mathcal{D}_\text{off}} \left[ G_t | s_t=s \right] \\
   G_t &= \sum_{i=0}^{T-t} \gamma^i r_{t+i}, \label{eq:mc_return}
\end{align}
where $T$ is the task horizon.
In practice, since we focus on continuous control problems, no two states in $\mathcal{D}_\text{off}$ are identical. This leads to a single trajectory return contributing to the Monte-Carlo value estimate. CalQL uses this estimate to replace the first term from \eqref{eq:cql} with
\begin{equation}
    \mathbb{E}_{s \sim \mathcal{D}_\text{off}, a \sim \pi_\text{off}(\cdot | s;\phi)}
            \left[
                \max (Q_\text{off}(s,a;\theta), V^{\pi_\beta}(s))
            \right].
\end{equation}
CQL and CalQL both learn $\pi_\text{off}$ with the SAC variant of \eqref{eq:actor_loss_general}.

\begin{figure*}[t]
\begin{minipage}{0.67\linewidth}
    \includegraphics[width=\linewidth]{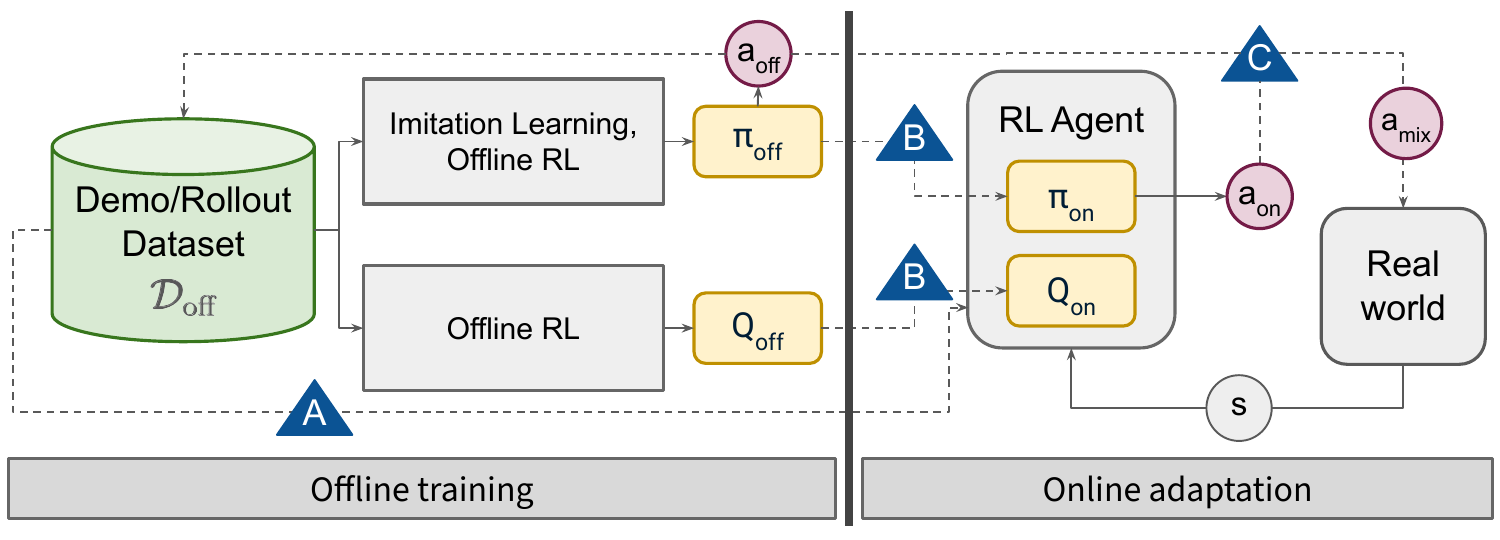}
    \caption{Three approaches for combining offline components with online RL. Strategy A samples data from $\mathcal{D}_\text{off}$ along with the online RL buffer. Strategy B uses pretrained $Q_\text{off}$ and $\pi_\text{off}$ and finetunes them with online experience. Strategy C uses actions from an offline policy as reference to generate a mixed action $a_\text{mix}$.}
    \label{fig:RainbowMain}
\end{minipage}%
\hfill%
\begin{minipage}{0.32\linewidth}
    \includegraphics[width=\linewidth,trim={2ex 0ex 2ex 0ex},clip]{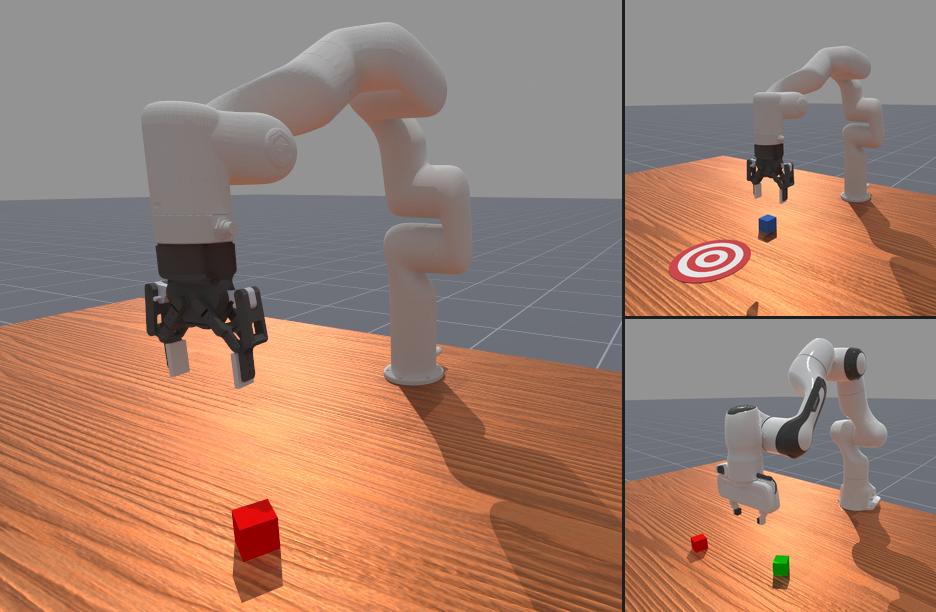}
    \caption{Tabletop manipulation tasks with Panda and xArm6 robots in the ManiSkill simulator \cite{maniskill3}.}
    \label{fig:maniskill_tasks}    
\end{minipage}
\end{figure*}

\section{Rainbow DemoRL}
\label{sec:rdemorl}

We identify three broad approaches for leveraging $\mathcal{D}_\text{off}$, $Q_\text{off}$, and $\pi_\text{off}$ to improve the sample efficiency of learning $Q_\text{on}$ and $\pi_\text{on}$ online. Strategy A uses the offline data $\mathcal{D}_\text{off}$ directly for online training. Strategy B first learns $Q_\text{off}$ and $\pi_\text{off}$ and uses them as initializations for $Q_\text{on}$ and $\pi_\text{on}$. Strategy C uses $\pi_\text{off}$ as a control prior to provide reference actions for the online policy.
Fig.~\ref{fig:RainbowMain} provides an overview of the components involved in each of the strategies. Below, we discuss implementing each strategy along with considerations for hybrid combinations.

\subsection{Strategy A: Using $\mathcal{D}_\text{off}$ directly}
\label{sec:pathA}
Off-policy actor-critic methods sample transitions from a replay buffer to update the policy and value parameters via \eqref{eq:critic_loss_general} and \eqref{eq:actor_loss_general}. A straightforward approach to incorporate an offline dataset is to sample some transitions from $\mathcal{D}_\text{off}$ along with the online buffer $\mathcal{D}_\text{on}$. We follow the recommendation from RLPD \cite{rlpd} to use 50/50 sampling from the offline dataset and online buffer, along with high UTD and a critic ensemble. Methods combining these elements are referred to as ``prefill'' methods in our experiments.

Instead of using transitions from $\mathcal{D}_\text{off}$ in the RL objectives, an alternative approach is to use the offline data to optimize an auxiliary BC loss \cite{td3bc}. This method adds a supervised term to the RL actor loss from \eqref{eq:actor_loss_general}, which is the BC loss from \eqref{eq:bc_stoch} or \eqref{eq:bc_det} for online SAC and TD3 respectively.

\subsection{Strategy B: Finetuning pretrained $Q_\text{off}$, $\pi_\text{off}$}
\label{sec:pathB}

Offline RL methods learn $Q_\text{off}$ and $\pi_\text{off}$ as discussed in Sec.~\ref{sec:offline_training}. We can use these as initializations for $\pi_\text{on}$ and $Q_\text{on}$, respectively. We follow prior work on offline RL finetuning \cite{wsrl} which finds that dropping conservative terms and using pure online RL objectives during finetuning leads to better asymptotic performance.

We also use a simple baseline, Monte-Carlo-Q (MCQ), which learns an offline Q-function as follows:
\begin{align*}
    L(\theta) = \mathbb{E}_{\tau \sim \mathcal{D}_\text{off}, t \in \{0,1,\dots,T\} } \left[( Q_\text{off}(s_t,a_t;\theta) - G_t)^2 \right],
\end{align*}
where $G_t$ is the Monte-Carlo return as defined in \eqref{eq:mc_return}. To ensure self-consistency of the learnt Q-function, with a small probability $\epsilon_b$, we use the TD error from \eqref{eq:critic_loss_general} to optimize the critic. In our experiments, whenever we initialize $Q_\text{on}$ using MCQ, we also initialize $\pi_\text{on}$ with a BC policy.

\subsection{Strategy C: Using $\pi_\text{off}$ for reference actions}
\label{sec:pathC}

Given a state $s$, we can generate both $a_\text{off} \sim \pi_\text{off}(\cdot | s)$ and $a_\text{on} \sim \pi_\text{on}(\cdot | s)$, and ``mix'' these actions in a meaningful way to help online learning. We consider the following three ideas for action mixing from IBRL \cite{ibrl}, CHEQ \cite{cheq} and Residual RL \cite{residualrl, residualrlrobot}.

\subsubsection{Higher Q-value \cite{ibrl}}
An action that maximizes the Q-values is chosen:
\begin{equation*}
    a_\text{mix} = \arg\max_{a \in \{a_\text{off}, a_\text{on}\}} Q_\text{on}(s, a)
\end{equation*}
and used for interaction with the environment as well as to compute the Q-target in \eqref{eq:q_target}.

Recent work \cite{valuelearningbottleneck} finds that offline RL methods learn a good value function $Q_\text{off}$ but struggle to extract a good policy. Hence, a natural hybrid method with this action mixing variant is to  leverage a pretrained $Q_\text{off}$ to initialize $Q_\text{on}$. This can enable effective action selection from the beginning of online training, without having to wait to first learn a good Q-function.

\subsubsection{Linear interpolation \cite{cheq}}
Here $a_\text{mix}$ is a linear interpolation of $a_\text{off}$ and $a_\text{on}$, with the online action given lower weight $\lambda \in [0,1]$ if the online critic is uncertain about its value estimate:
\begin{equation*}
    a_\text{mix} = (1 - \lambda) a_\text{off} + \lambda a_\text{on}.
\end{equation*}
The uncertainty is measured by the variance of the Q-value predictions across the Q-ensemble:
\begin{equation*}
    \lambda \propto \frac{1}{Var_i[Q_{\text{on},i}(s, a_\text{on})]}.
\end{equation*}
This approach also requires the mixing weight to be observable by RL to allow it to adapt its actions correctly.
Hence, we use a modified state space $\mathcal{S}' \subseteq \mathbb{R}^{n+1}$ with state vectors $s' = [s^\top, \lambda]^\top$ for all CHEQ-based methods.

\subsubsection{Residual RL \cite{residualrl}}
Residual Policy Learning (RPL) considers learning $a_{\text{on}}$ as a residual corrective action on top of $a_\text{off}$; that is, $a_{\text{mix}} = a_{\text{off}} + a_{\text{on}}$. Consistent with recent work \cite{residualoffpolicyrl, maniptrans}, the residual actor is conditioned on the offline action, and its final layer is zero-initialized to ensure $a_{\text{mix}} \approx a_{\text{off}}$ during initial interaction. The online learning objectives remain identical to the standard setting in Sec.~\ref{sec:online_rl} with $a_{\text{on}}$ simply replaced by $a_{\text{mix}}$.

\begin{figure}[!t]
    \vspace*{3mm}
    \begin{subfigure}[b]{0.5\linewidth}
        \includegraphics[width=\linewidth]{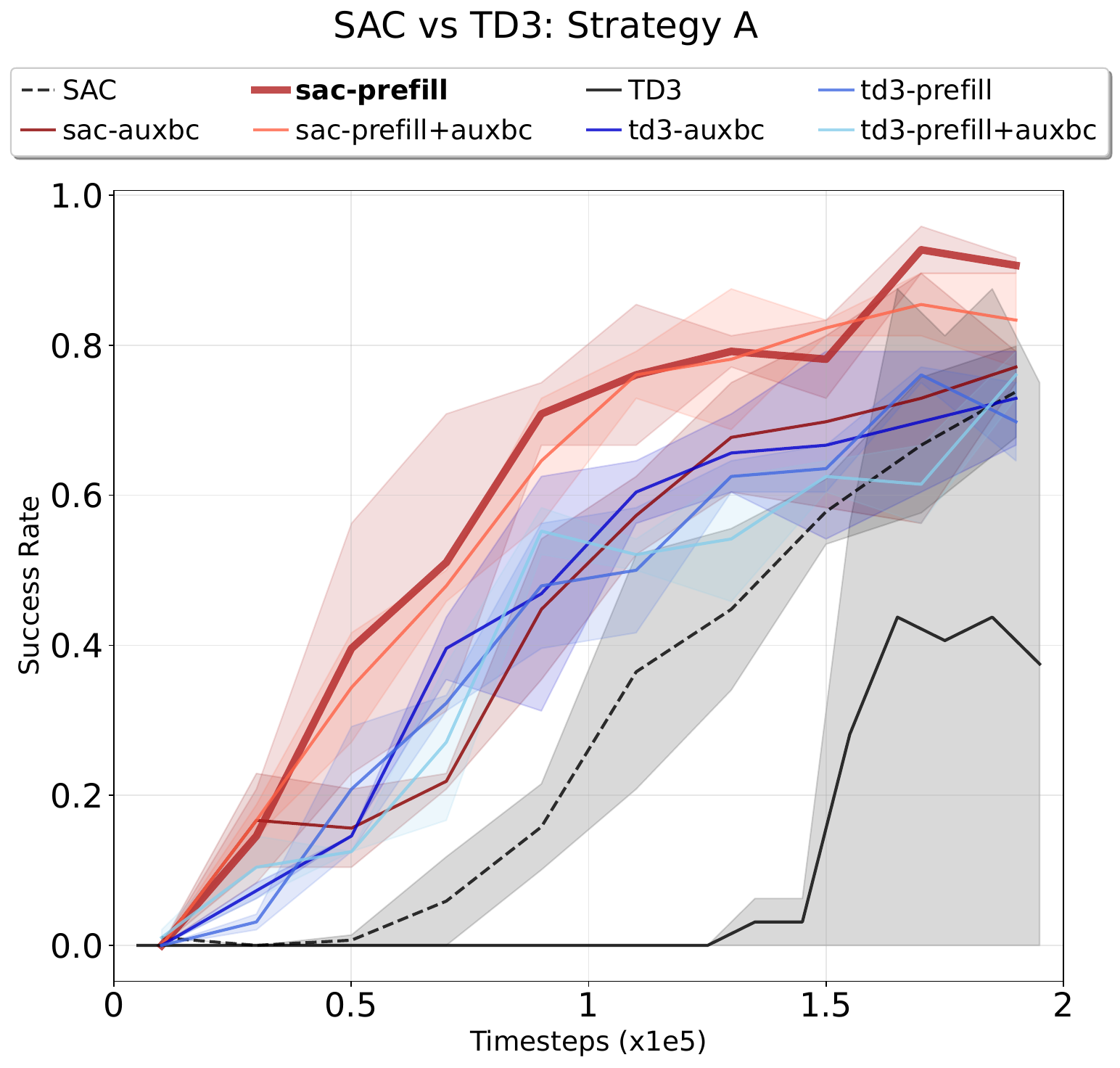}
    \end{subfigure}
    \hfill
    \begin{subfigure}[b]{0.48\linewidth}
        \includegraphics[width=\linewidth]{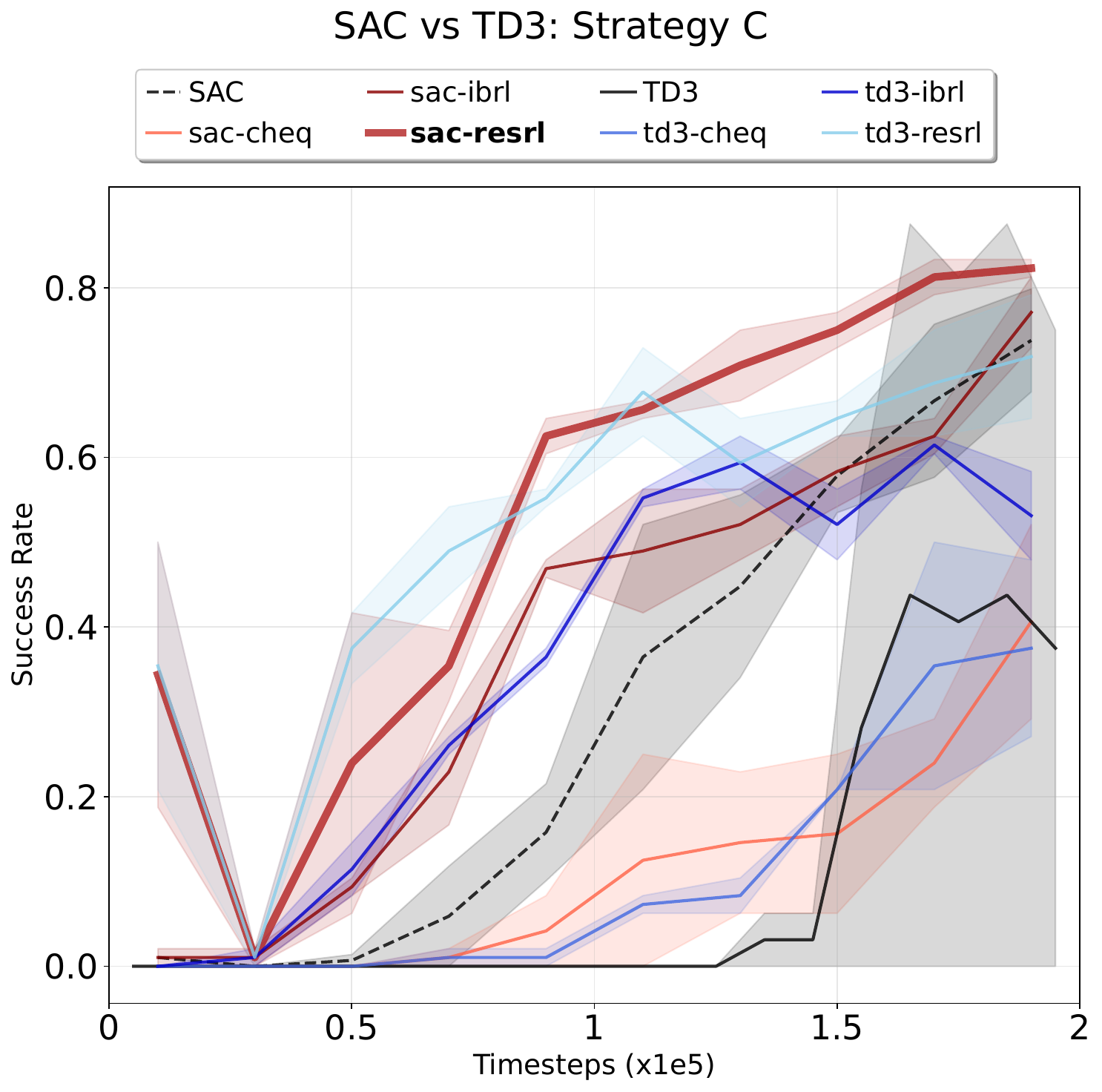}
    \end{subfigure}
    \caption{Comparison of SAC vs TD3 success rates for Strategy A and C approaches.}
    \label{fig:path_A_C_sac_td3}
\end{figure}

\section{Experiments}
\label{sec:experiments}

We evaluate all combinations of strategies described in Sec.~\ref{sec:rdemorl} against a Soft Actor-Critic (SAC) \cite{sac} baseline across five scenarios\footnote{A full set of results is available on our public wandb dashboard at \url{https://wandb.ai/ucsd_erl/rainbow-demorl-final}}. These scenarios consist of the PickCube, PushCube, and StackCube tasks from ManiSkill \cite{maniskill3} on uFactory xArm6 and Franka Emika Panda robots (we exclude the StackCube-xArm6 configuration, where the SAC baseline failed to learn within a budget of 1M steps). The setup for these tasks is visualized in Fig.~\ref{fig:maniskill_tasks}. The state $s$ is composed of robot proprioception and distances between the end effector and relevant objects, while the actions $a$ are joint velocities. We use dense rewards from Maniskill for each task. Our demonstrations consist of 5{,}000–10{,}000 trajectories per task, collected from the replay buffer of an RL expert trained for over 1M steps. To ensure high-quality data, we filter for episodes that achieve at least 90\% of the peak return observed during the expert's training.

To evaluate both learning speed and asymptotic performance, we compute the area-under-curve (AUC) of the success rate with respect to online interaction steps. Our evaluation proceeds in three stages. First, for each environment and robot configuration, we calculate the percentage improvement of an algorithm $x$ over the SAC baseline as $(AUC_x - AUC_{\text{SAC}}) / AUC_{\text{SAC}}$, averaged across three random seeds. Second, we apply a sign-preserving normalization where positive and negative improvements are scaled to $[0, 1]$ and $[-1, 0]$ respectively, anchored by the best and worst performing algorithms in each setting. Finally, we average these normalized values across all experiment settings to produce the Sample Efficiency Improvement (SEI) score for algorithm $x$. This metric provides a balanced comparison of algorithmic performance that is robust to task-specific variance and prevents outliers in easier tasks from skewing the aggregate results.

\subsection{Base RL algorithm}

We compare SAC \cite{sac} and TD3 \cite{td3} variants of single strategy methods to fix the base online RL algorithm. Methods employing strategy B naturally work better with SAC since conservative offline RL algorithms like CQL and CalQL learn a stochastic policy. We also observe that for strategies A and C, SAC variants have lower variance across random seeds and often outperform their TD3 counterparts, as seen in the comparison of these algorithms for PickCube in Fig.~\ref{fig:path_A_C_sac_td3}. Hence, in the remainder of the experiments, we use SAC as the base RL algorithm.

\begin{figure}[t!]
    \vspace*{3mm}
    \centering
    \begin{subfigure}{0.9\columnwidth}
        \centering
        \includegraphics[width=\linewidth,trim={0 6ex 0 9ex},clip]{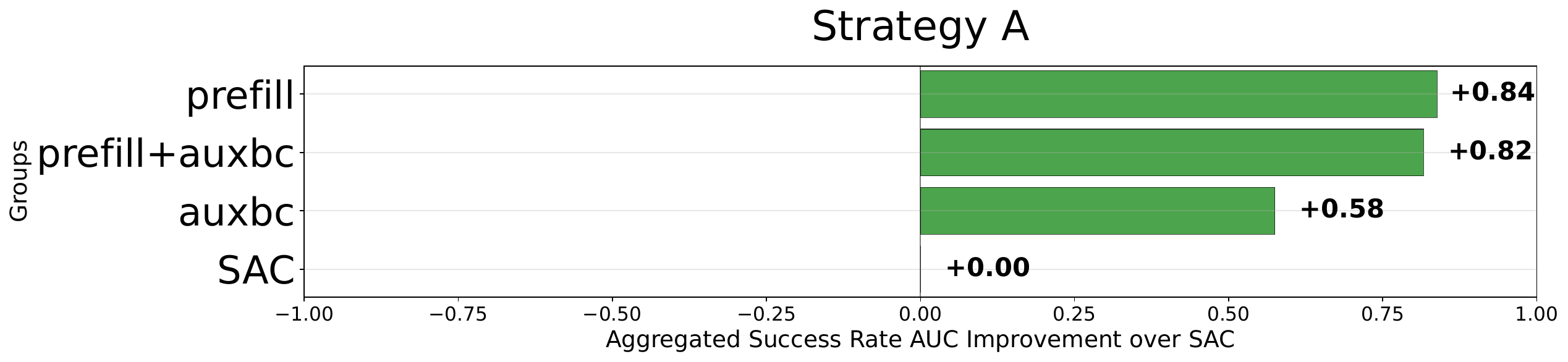}
        \caption{Comparison of SEI scores for strategy A approaches using $\mathcal{D}_\text{off}$ with online RL (SAC).}
        \label{fig:pathA}
    \end{subfigure}\\[0.5em]
    \begin{subfigure}{0.9\columnwidth}
        \centering
        \includegraphics[width=\linewidth,trim={0 6ex 0 9ex},clip]{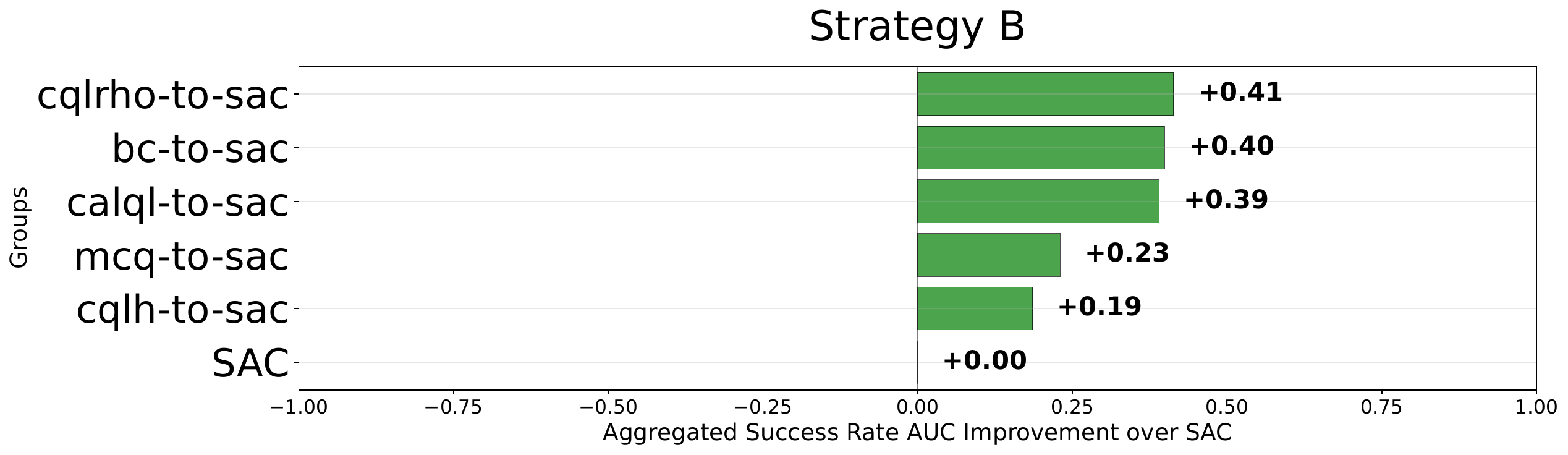}
        \caption{Comparison of SEI scores for strategy B approaches finetuning pretrained policies and value functions.}
        \label{fig:pathB}
    \end{subfigure}\\[0.5em]
    \begin{subfigure}{0.9\columnwidth}
        \centering
        \includegraphics[width=\linewidth,trim={0 6ex 0 9ex},clip]{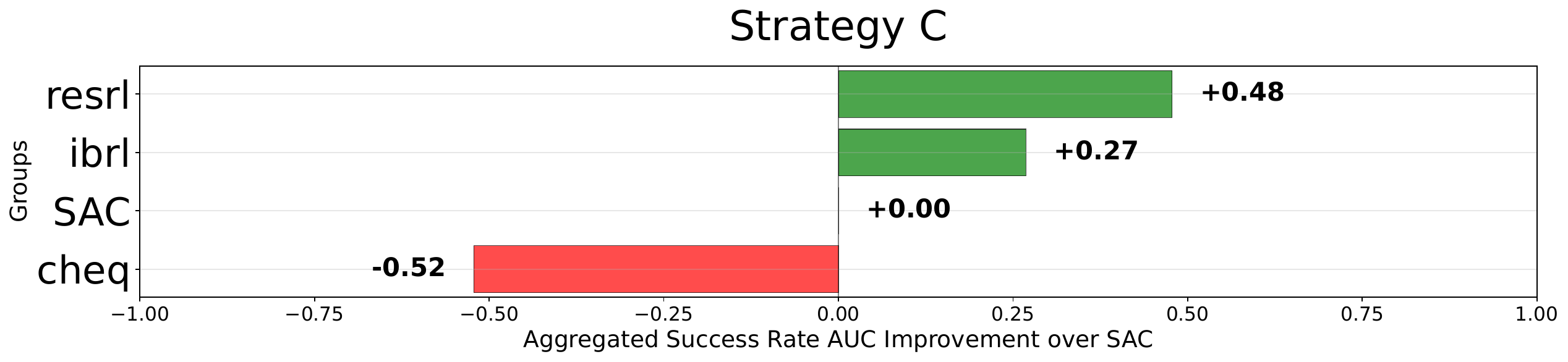}
        \caption{Comparison of SEI scores for strategy C approaches using $\pi_\text{off}$ for reference actions.}
        \label{fig:pathC}
    \end{subfigure}
    \caption{Comparing Sample Efficiency Improvement (SEI) scores for single strategy approaches.}
    \label{fig:single_strategies}
\end{figure}

\begin{figure*}[t!]
    \centering
    \begin{subfigure}{.48\linewidth}
        \centering
        \includegraphics[width=\linewidth,trim={0 7ex 0 9ex},clip]{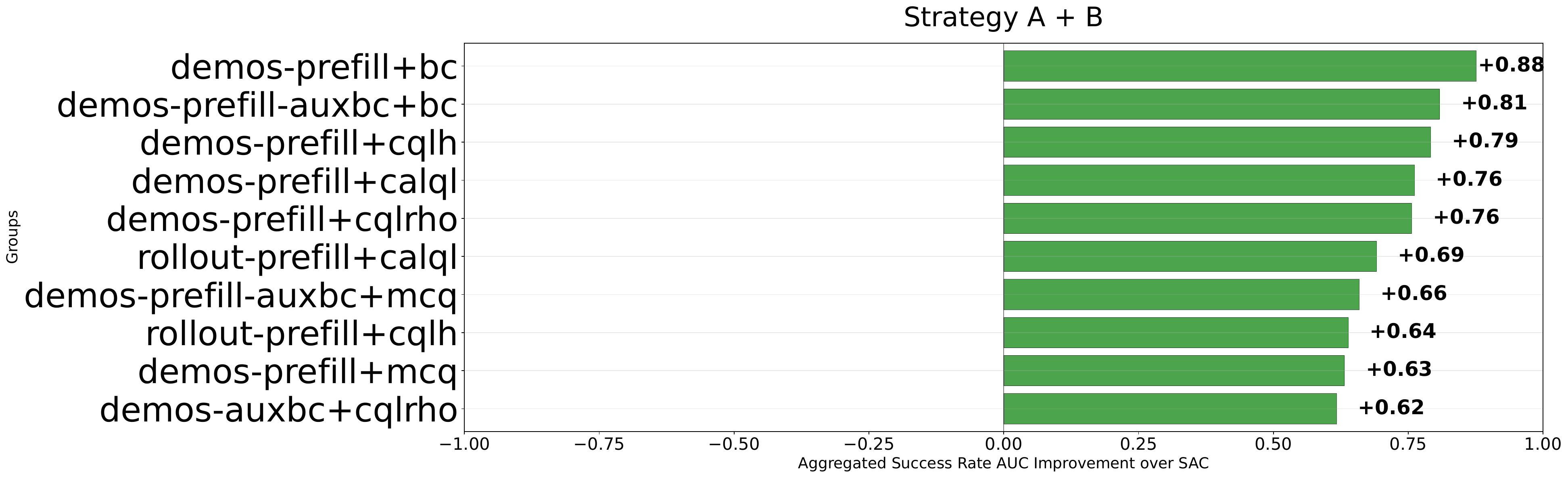}
        \caption{Comparison of SEI scores for the top ten hybrid variants combining strategies A and B.}
        \label{fig:pathsAB}
    \end{subfigure}%
    \hfill
    \begin{subfigure}{.48\linewidth}
        \vspace*{3mm}
        \centering
        \includegraphics[width=\linewidth,trim={0 7ex 0 9ex},clip]{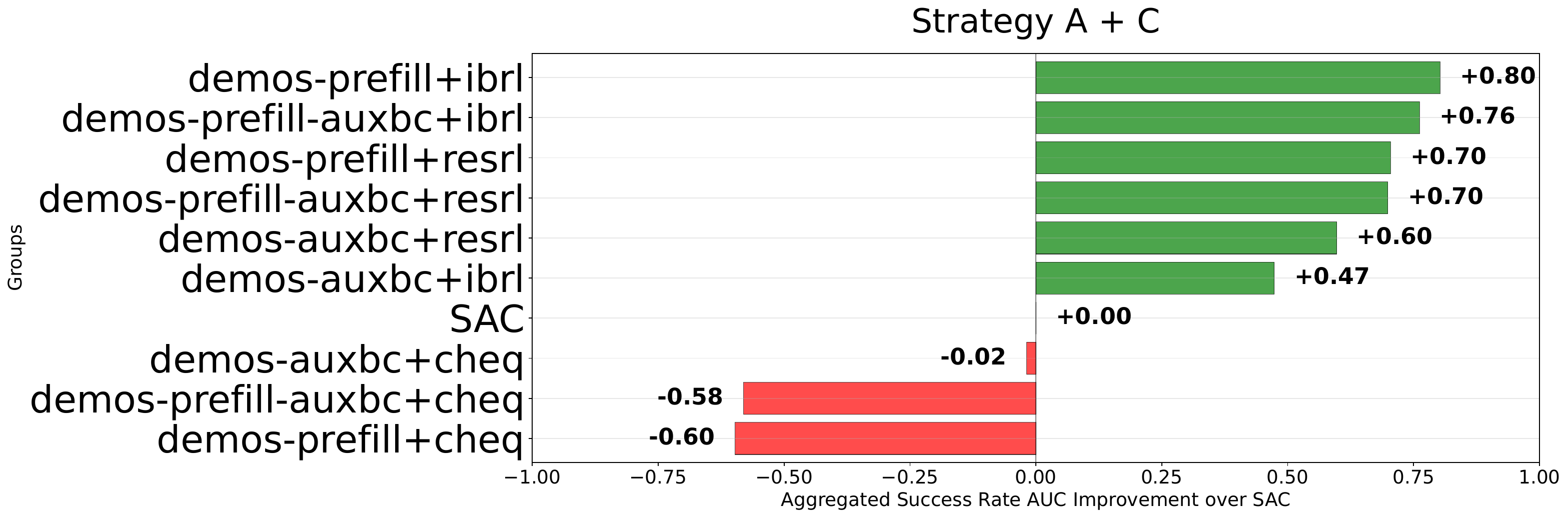}
        \caption{Comparison of SEI scores for the top ten hybrid variants combining strategies A and C.}
        \label{fig:pathsAC}
    \end{subfigure}

    \vspace{0.5em}

    \begin{subfigure}{.48\linewidth}
        \centering
        \includegraphics[width=\linewidth,trim={0 6ex 0 9ex},clip]{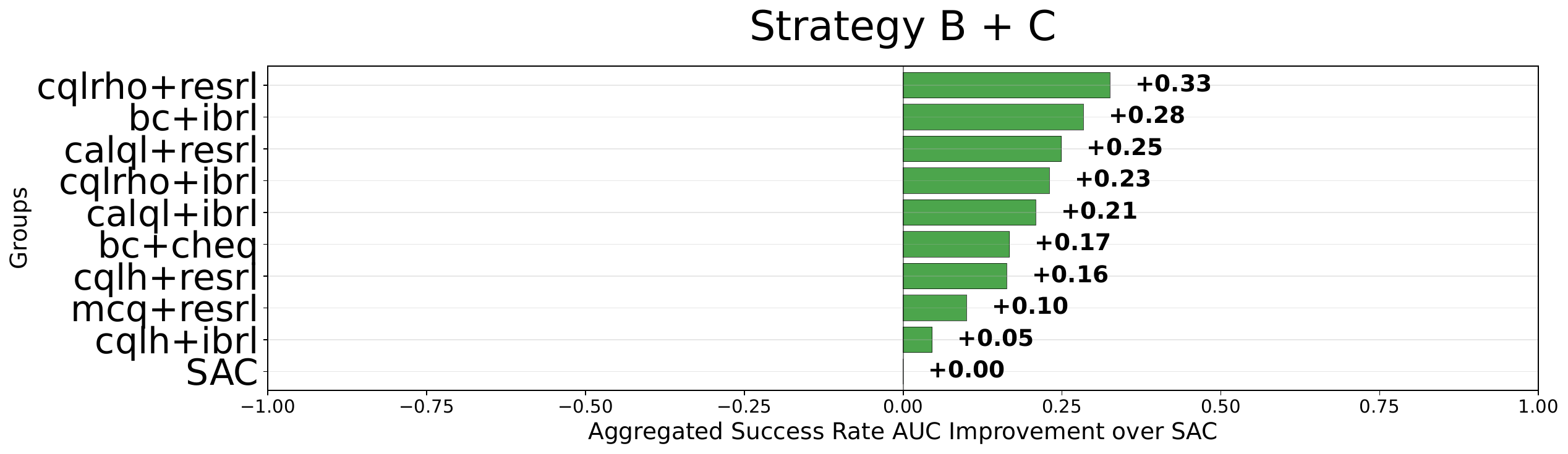}
        \caption{Comparison of SEI scores for the top ten hybrid variants combining strategies B and C.}
        \label{fig:pathsBC}
    \end{subfigure}%
    \hfill
    \begin{subfigure}{.48\linewidth}
        \centering
        \includegraphics[width=\linewidth,trim={0 6ex 0 9ex},clip]{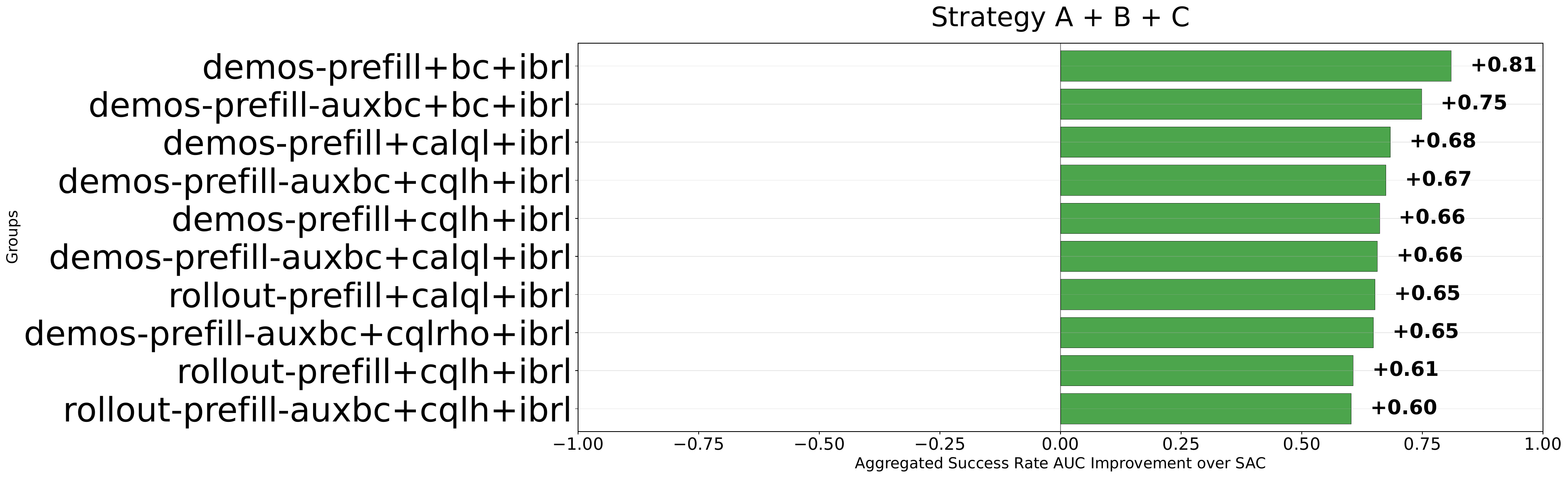}
        \caption{Comparison of SEI scores for the top ten hybrid variants combining strategies A, B and C.}
        \label{fig:pathsABC}
    \end{subfigure}

    \caption{Comparison of Sample Efficiency Improvement (SEI) scores for the top ten hybrid variants across different strategy combinations.}
    \label{fig:paths_all}
\end{figure*}

\subsection{Single strategy}

\subsubsection{Only A}

Fig.~\ref{fig:pathA} compares the performance of strategy A variants with BC auxiliary loss \cite{td3bc} and prefilling the replay buffer with offline data \cite{rlpd}. While all variants here are more efficient than the baseline, the fastest approach is the one that uses buffer prefilling like RLPD \cite{rlpd}, with the approach combining prefill and auxiliary BC loss following closely behind.

\subsubsection{Only B}
\label{sec:eval_onlyb}
Fig.~\ref{fig:pathB} compares the performance of strategy B variants that finetune pretrained value and policy models. We first consider the case of pretraining an offline BC policy and starting with an untrained critic. Next, we consider jointly pretrained actors and critics with offline RL methods CQL and CalQL, and finally with MCQ described in Sec.~\ref{sec:pathB}. For CQL, there are two ways of implementing the first term from \eqref{eq:cql}, corresponding to CQL-H and CQL-$\rho$ \cite{cql}. 

While all finetuning strategies improve online sample efficiency over SAC, pretraining with CQL-$\rho$ emerges as the fastest algorithm in this category, with simple BC policy pretraining with an untrained critic as a close runner-up.
Despite the gains from finetuning, no strategy B approach learns as fast as Strategy A methods.

\subsubsection{Only C}

Fig.~\ref{fig:pathC} evaluates the performance of strategy C variants, which compare the Q-values of the online and offline actions (IBRL \cite{ibrl}), use linear interpolation between $a_\text{off}$ and $a_\text{on}$ (CHEQ \cite{cheq}), or learn $a_\text{on}$ as a residual on top of $a_\text{off}$ (RPL \cite{residualrl}). All methods except CHEQ lead to improvements in online sample efficiency. RPL achieves the fastest learning and typically also has the best initial performance
since it starts with a zero-initialized online policy leading to $a_\text{mix} \approx a_\text{off}$. Overall, strategy C methods have higher variance in performance, and their sample efficiency gains are highly dependent on the specific action mixing variant.

\subsection{Hybrid strategies}
While single-strategy methods provide significant gains, we test if integrating multiple approaches can yield further improvements in sample efficiency.
Certain hybrid methods benefit from specific modifications to the individual strategies to maximize the performance. For A+B hybrids that both prefill the replay buffer
and use a pretrained policy $\pi_\text{off}$, a key consideration is the data source used for pre-filling the buffer.
WSRL \cite{wsrl} finds that prefilling
with rollouts from $\pi_\text{off}$ instead of the original offline dataset improves sample efficiency and final agent performance. 

For B+C hybrids that use pretrained value and policy along with CHEQ for action mixing, we need to account for the updated state space $\mathcal{S}' \subseteq \mathbb{R}^{n+1}$, which necessitates larger input layers for the Q and policy networks than the ones from $Q_\text{off}$ and $\pi_\text{off}$. We handle these differences by partially loading the pretrained weights onto $Q_\text{on}$ and $\pi_\text{on}$ for the first $n$ neurons in the input layer, and randomly initializing weights for the new input neuron corresponding to $\lambda$.

The relative effectiveness of hybrid approaches is summarized in the following comparisons from Fig~\ref{fig:paths_all}, which highlights the top ten variants identified for each strategy combination.
A+B hybrids learn significantly faster than the baseline. However, the performance of the best A+B hybrids is similar. The combination with demonstration prefill and simple BC initialization achieves better performance than conservative offline RL methods, as seen in Fig.~\ref{fig:pathsAB}. 
A+C hybrids are compared in Fig.~\ref{fig:pathsAC}, which shows that combinations using IBRL and residual RL learn significantly faster than those using CHEQ, which at best matches the baseline performance. The fastest approach here is IBRL with a replay buffer prefilled with demonstration data. 
Next, we see from Fig.~\ref{fig:pathsBC} that B+C hybrids achieve only moderate sample-efficiency gains over the baseline. This indicates that strategy A of using samples from the offline dataset directly is critical to achieving sample-efficient online RL.
Finally, methods combining all three strategies also improve on the baseline with the prefill and IBRL components commonly showing up among the best performers as seen in Fig.~\ref{fig:pathsABC}.

\begin{figure}[t]
    \centering
    \vspace{-7pt}
    \includegraphics[width=\linewidth,trim={0 6ex 0 9ex},clip]{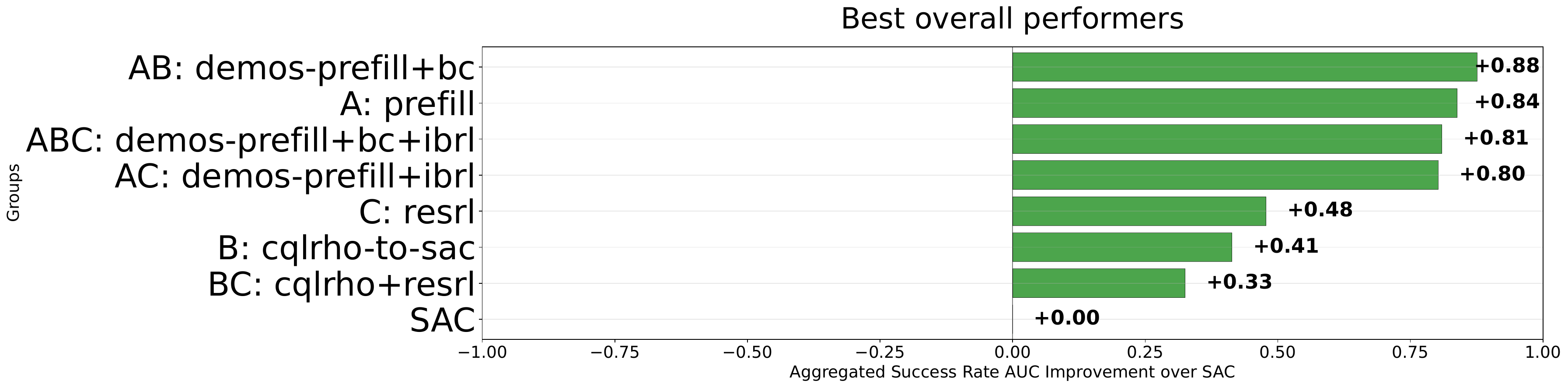}
    \caption{Comparison of Sample Efficiency Improvement scores for the best performing variant of each strategy combination.}
    \vspace{-7pt}
    \label{fig:best_overall}
\end{figure}

\begin{figure*}[p]
    \vspace*{2mm}
    \centering
    \includegraphics[width=0.9\linewidth,trim={0 6ex 0 20ex},clip]{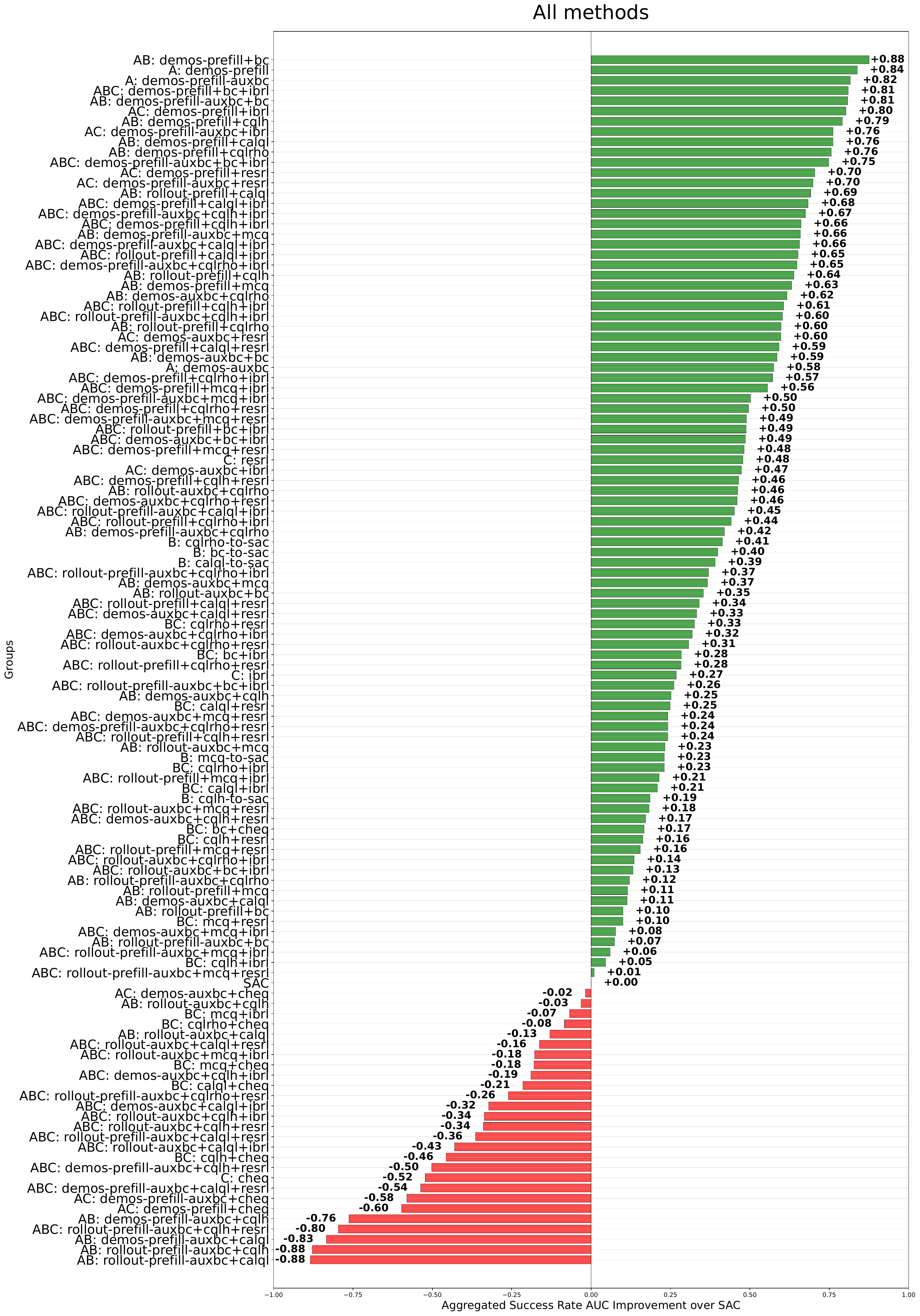}
    \caption{Comparison of Sample Efficiency Improvement (SEI) scores for all methods, aggregated across the four evaluated task-robot configurations. Higher SEI scores indicate faster learning speed and final performance relative to the SAC baseline (SEI = 0). Component abbreviations are defined as follows: \textit{demos/rollouts} indicates the data source; \textit{prefill/auxbc} denotes the Strategy A direct data usage method; \textit{bc/mcq/calql/cqlrho/cqlh} specifies the Strategy B pretraining method; and \textit{resrl/ibrl/cheq} represents the Strategy C action mixing mechanism.}
    \label{fig:auc_all}
\end{figure*}

We perform a comparison across the best methods for each strategy combination in Fig.~\ref{fig:best_overall}. We note that the hybrid methods employing strategy A outperform the rest. Finally, an extensive comparison of all methods considered in the paper is presented in Fig.~\ref{fig:auc_all}. We find that the A+B approach that uses a replay buffer prefilled with demonstration data along with a simple pretrained BC policy improves the sample efficiency of online RL the most. Other hybrids using buffer prefill and an auxiliary BC loss follow closely behind in performance.

\begin{figure*}[t!]
    \centering
    \includegraphics[width=0.98\linewidth,trim={0 0 0 8ex},clip]{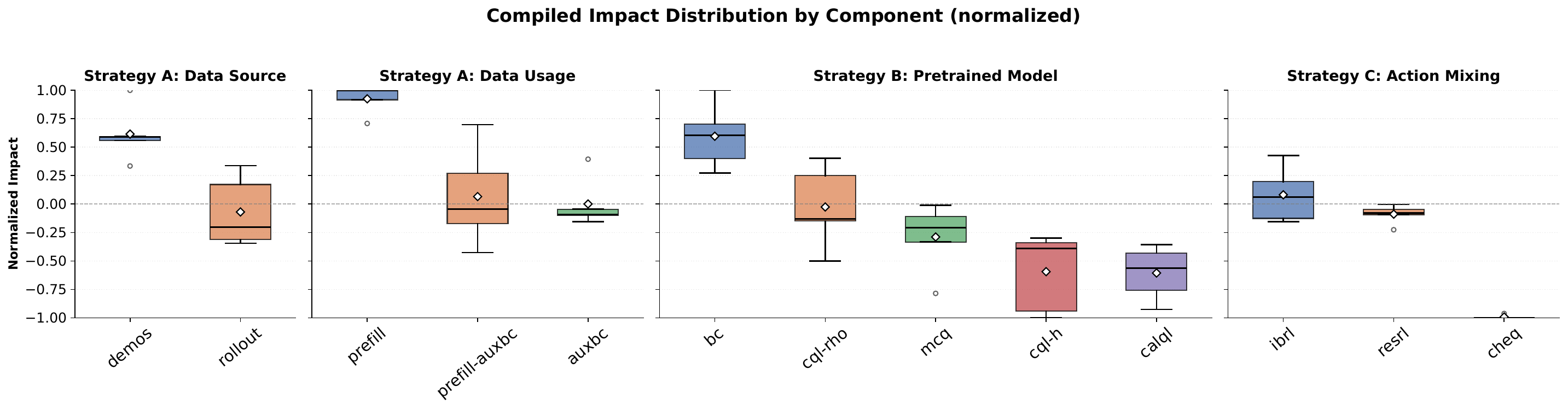}
    \caption{Distribution of normalized impact scores for individual algorithmic components across all task and robot configurations. Scores are independently normalized to $[-1, 1]$ per configuration. Positive values indicate that adding the component improved sample efficiency compared to matched baselines, while negative values indicate a degradation. The spread of the whiskers illustrates each component's sensitivity to different task difficulties.}
    \label{fig:impact_whisker}
\end{figure*}

\begin{figure*}[t]
    \centering
    \includegraphics[width=0.98\linewidth,trim={0 5ex 0 8ex},clip]{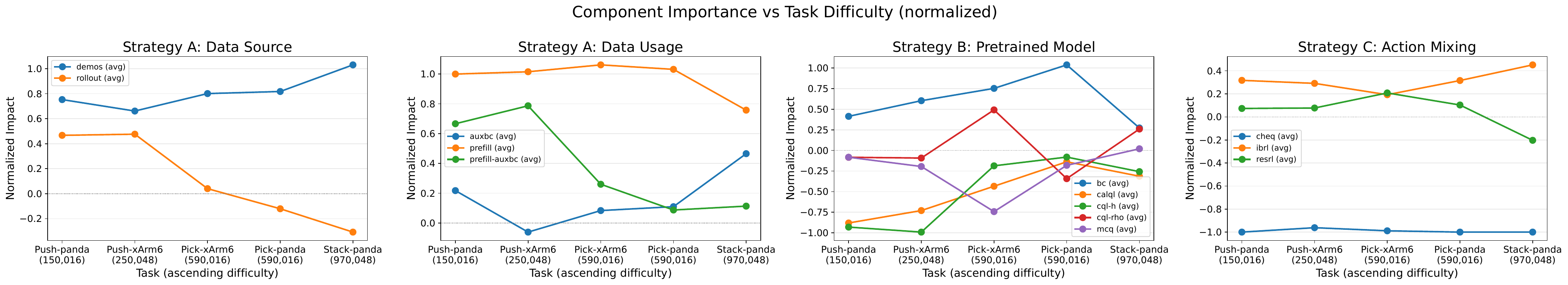}
    \caption{Normalized impact of individual components as a function of task difficulty. Tasks are ranked from easiest to hardest, defined by the number of steps required for SAC to reach peak success (values shown in parentheses below each task). Results are grouped by strategy to highlight how the utility of each algorithmic choice shifts in more complex settings.}
    \label{fig:impact_difficulty}
\end{figure*}

\subsection{Per-component impact}
To isolate the marginal contribution of individual algorithmic choices in hybrid algorithms, we conduct a paired ablation analysis. For a given environment and robot configuration, we identify all algorithm pairs that differ solely by the presence of a specific component $x$. The relative performance gain yielded by $x$ is calculated as $AUC_{\text{with } x} - AUC_{\text{without } x}$. We use the difference rather than a ratio to ensure the impact score remains robust across diverse baseline performance levels ($AUC_{\text{without } x}$) and to avoid over-weighting improvements over weak ablation baselines. We average this value across all valid algorithm pairs to determine the raw impact of component $x$ in that specific setting. To account for the inherent variance in task difficulty, these per-setting impact scores undergo the same sign-preserving normalization used for our SEI metric. Positive and negative impacts are scaled independently to $[0, 1]$ and $[-1, 0]$, anchored by the highest and lowest performing components within each setting. 

To capture both the average impact and the cross-task variance of each design choice, we visualize the distribution of these normalized impact scores across all experimental settings in Fig.~\ref{fig:impact_whisker}. 
Notably, while certain value function pretraining methods like CQL-$\rho$ show strong standalone performance in isolation (Sec.~\ref{sec:eval_onlyb}), simple BC policy pretraining is the only Strategy B component that consistently yields positive marginal gains across the full diversity of hybrid configurations. This suggests that the sample overhead of recalibrating a conservative offline critic often outweighs its initialization benefits when combined with other acceleration strategies.
Conversely, direct data usage via buffer prefilling like RLPD proves to be a universally effective component, consistently improving sample efficiency across all environments.

Finally, we evaluate the relative impact of each component as a function of task difficulty, quantified by the number of online interaction steps required for the SAC baseline to reach its peak success rate (Fig.~\ref{fig:impact_difficulty}). For strategy A, we observe a strong negative correlation between task difficulty and the utility of policy rollouts as a data source. This is expected, as the behavioral quality of a pre-trained policy's rollouts naturally degrades in more complex environments. In addition, as task complexity scales, the relative impact of buffer prefilling diminishes, while that of an auxiliary BC loss increases. Among strategy B and C components, the most pronounced trends emerge from the offline RL methods CalQL and CQL-H, both of which exhibit an increasingly positive impact on sample efficiency as tasks become more challenging.

\section{Conclusion}
We presented Rainbow-DemoRL, a systematic categorization and large empirical evaluation of demonstration-augmented online reinforcement learning. We organized existing methods into three core strategies of direct data usage, model fine-tuning, and reference action mixing, and evaluated both their isolated and combined impacts on online sample efficiency. Our ablation study led to the following practical recommendation for robotic manipulation using reinforcement learning: complex offline RL value pretraining often slows down online adaptation due to the sample overhead required for critic recalibration. Instead, the most performant and sample-efficient approaches employ replay buffer prefilling \cite{rlpd} with a simple behavior cloning policy initialization and an untrained critic. We hope the comprehensive analysis presented in this work serves as a valuable reference for researchers seeking to deploy manipulation policies that adapt rapidly during online interaction with the world.

\bibliographystyle{ieeetr}
\bibliography{references}

\appendix
\label{sec:appendix_hyperparams}

All experiments use the hyperparameters listed in Table \ref{tab:hyperparameters}.

\begin{table}[h]
\centering
\caption{Hyperparameters}
\label{tab:hyperparameters}
\begin{tabular}{@{}ll@{}}
\toprule
\textbf{Hyperparameter} & \textbf{Value} \\ \midrule
\multicolumn{2}{c}{\textit{Online RL Hyperparameters}} \\ \midrule
Horizon ($h$) & 3 \\
Discount factor ($\gamma$) & 0.8 \\
Target smoothing ($\tau$) & 0.01 \\
Batch size & 1024 \\
Learning rate & $3 \times 10^{-4}$ \\
Update-to-data ratio & 1 / 5 (for prefill methods) \\ \midrule
\multicolumn{2}{c}{\textit{Neural Network Architecture}} \\ \midrule
MLP hidden dimension & 256 \\
Actor hidden layers & 3 \\
Critic hidden layers & 3 \\
Number of critics & 2 / 5 (for prefill methods) \\ \midrule
\multicolumn{2}{c}{\textit{Algorithm-Specific Parameters}} \\ \midrule
\textbf{TD3} & \\
\quad Target policy noise & 0.1 \\
\quad Noise clip & 0.5 \\
\quad Exploration noise & 0.1 \\
\textbf{SAC} & \\
\quad Entropy coeff. ($\alpha$) & 0.2 \\
\textbf{CQL} & \\
\quad Actor learning rate & $3 \times 10^{-5}$ \\
\quad CQL weight & 5.0 \\
\textbf{MCQ} & \\
\quad Bootstrap epsilon ($\epsilon_b$) & 0.1 \\
\textbf{CHEQ} & \\
\quad Uncertainty bounds ($u$) & [0.15, 0.275] \\
\quad Lambda bounds ($\lambda$) & [0.2, 1.0] \\
\textbf{Residual RL} & \\
\quad Critic burn-in steps & 20,000 \\
\bottomrule
\end{tabular}
\end{table}

\end{document}